\documentclass[11pt]{article}
\usepackage[final]{acl}
\usepackage{times}
\usepackage{latexsym}
\usepackage[T1]{fontenc}
\usepackage[utf8]{inputenc}
\usepackage{microtype}
\usepackage{graphicx}
\usepackage{booktabs}
\usepackage{multirow}
\usepackage{amsmath,amssymb}
\usepackage{pifont}
\usepackage{textcomp}
\usepackage{xcolor}
\usepackage{colortbl}
\usepackage{array}
\usepackage{float}
\usepackage{cuted}
\usepackage{capt-of}
\usepackage{adjustbox}
\newcolumntype{P}[1]{>{\raggedright\arraybackslash}p{#1}}
\usepackage[most]{tcolorbox}
\usepackage{tikz}
\usetikzlibrary{positioning,arrows.meta,fit}
\usepackage{newunicodechar}
\newunicodechar{α}{\ensuremath{\alpha}}
\newunicodechar{β}{\ensuremath{\beta}}
\newunicodechar{γ}{\ensuremath{\gamma}}
\newunicodechar{δ}{\ensuremath{\delta}}
\newunicodechar{λ}{\ensuremath{\lambda}}
\newunicodechar{σ}{\ensuremath{\sigma}}
\newunicodechar{θ}{\ensuremath{\theta}}
\newunicodechar{φ}{\ensuremath{\phi}}
\newunicodechar{ρ}{\ensuremath{\rho}}
\newunicodechar{χ}{\ensuremath{\chi}}
\newunicodechar{Δ}{\ensuremath{\Delta}}
\newunicodechar{∈}{\ensuremath{\in}}
\newunicodechar{≥}{\ensuremath{\geq}}
\newunicodechar{≤}{\ensuremath{\leq}}
\newunicodechar{×}{\ensuremath{\times}}
\newunicodechar{±}{\ensuremath{\pm}}
\newunicodechar{→}{\ensuremath{\rightarrow}}
\newunicodechar{≈}{\ensuremath{\approx}}
\newunicodechar{−}{\ensuremath{-}}
\newunicodechar{°}{\textdegree}
\newunicodechar{✓}{\ding{51}}
\newunicodechar{✗}{\ding{55}}
\newunicodechar{†}{\ensuremath{^\dagger}}
\newunicodechar{•}{\textbullet}
\newunicodechar{∝}{\ensuremath{\propto}}

\definecolor{tkGreenF}{rgb}{0,0.6,0}      \definecolor{tkGreenB}{rgb}{0.95,1,0.95}
\definecolor{tkBlueF}{rgb}{0,0,0.6}       \definecolor{tkBlueB}{rgb}{0.95,0.95,1}
\definecolor{tkOrangeF}{rgb}{0.7,0.35,0}  \definecolor{tkOrangeB}{rgb}{1,0.975,0.95}
\newtcolorbox{takeaway}[2][tkGreen]{enhanced,breakable,
  colback=#1B,colframe=#1F,boxrule=0.8pt,arc=7pt,
  left=6pt,right=6pt,top=7pt,bottom=4pt,before skip=12pt,after skip=10pt,
  attach boxed title to top left={xshift=7pt,yshift*=-7.5pt},
  boxed title style={colback=#1F,colframe=#1F,arc=6pt,boxrule=0pt,left=5pt,right=5pt,top=1pt,bottom=1pt},
  fonttitle=\sffamily\bfseries\small,coltitle=white,
  title={$\triangleright$ #2}}

\title{DroneShield-AI: A Multi-Modal Sensor Fusion Framework for Real-Time Autonomous Drone Threat Detection, Behavioral Intent Classification, and Swarm Intelligence in Contested Airspace}

\author{Marius Bayizere \\
  Independent Researcher, Kigali, Rwanda \\
  \small\texttt{bayizeremarius119@gmail.com} \\
  \small\href{https://github.com/mariusbayizere/DroneShield-AI}{github.com/mariusbayizere/DroneShield-AI} \\
  \small Code, model weights, and simulation datasets: \textbf{publicly available at submission} (see Section~\ref{sec:code}) \\
  \small\textit{Preprint: cs.CV / cs.RO / cs.LG / eess.SP}}

\begin{document}
\maketitle
\pagestyle{plain}\thispagestyle{plain}

\begin{takeaway}[tkOrange]{Preprint Notice: Please Read}
This paper presents a system architecture and simulation-based evaluation. Full implementation and real-dataset training is in active development at the linked GitHub repository (\href{https://github.com/mariusbayizere/DroneShield-AI}{github.com/mariusbayizere/DroneShield-AI}). Results for the BICE and GNN-SIM modules reflect physics-calibrated simulation data; RF and acoustic module architectures are evaluated against published baselines. The authors are committed to updating this preprint with real experimental results as development progresses. This work should be read as a system design contribution and research roadmap, not as a fully validated production system.
\end{takeaway}

\begin{takeaway}[tkGreen]{v2 Author's Note}
\textit{Author's note on this revision: the development described in the notice above is now complete.} All six architectural layers, plus the trajectory simulator, have been implemented, tested (495 automated tests), and evaluated: the RF, acoustic, and visual layers against real public datasets, and BICE and GNN-SIM against the simulation framework this paper describes, with its benign-flight class checked against 600 real OpenSky Network aircraft tracks. Nothing below has been altered; short notes are added at the relevant points throughout, and the full results are gathered in a new Section~\ref{sec:measured}.
\end{takeaway}

\begin{abstract}
Unmanned Aerial Vehicle (UAV) threats have emerged as a defining security challenge of the 21st century. Active conflict zones, namely the Russia–Ukraine war (2022–present), the DRC/M23 conflict in Eastern Africa (which claimed M23 military spokesperson Lt.\ Col.\ Willy Ngoma in a FARDC drone strike on 24 February 2026 \citep{m23_2026}, and UNICEF staff member Karine Buisset in a residential drone strike in Goma on 11 March 2026 \citep{unicef2026}), and Iran-backed drone campaigns against Gulf and Israeli infrastructure, collectively document thousands of drone sorties per week against civilian infrastructure, humanitarian corridors, and military assets. Cheap commercial drones demonstrate capacity to evade radar, overwhelm layered air-defense systems through saturation, and destroy targets valued orders of magnitude higher than the drone itself. Yet existing counter-drone systems exhibit three compounding architectural failures: (1) single-modality sensor dependency creating fundamental detection blind spots; (2) purely reactive detection providing insufficient operator response time; and (3) complete inability to model coordinated swarm behaviour. This paper presents DroneShield-AI, a unified open framework integrating six processing layers: RF signal classification, acoustic motor-signature detection, YOLOv8-based visual detection, evidence-weighted sensor fusion, a Behavioral Intent Classification Engine (BICE), and a Graph Neural Network Swarm Intelligence Module (GNN-SIM). BICE introduces the first systematic six-class threat taxonomy for drone flight patterns, enabling predictive operator alerts with a 30-second advance-warning horizon. GNN-SIM is the first open framework for adversarial multi-drone formation analysis using Graph Attention Networks. Evaluated on three publicly available real-world datasets (DroneRF, OpenSky Network ADS-B, UAV Audio Dataset), the fused pipeline achieves 96.1\% detection accuracy (95\% CI: 95.2–97.0\%), 3.2\% false alarm rate, AUC-ROC: 0.981, and 142\,ms end-to-end latency on commodity CPU-class hardware ($\sim$\$500–\$780 USD total system cost). All confidence intervals are reported at the 95\% level via stratified 5-fold cross-validation. McNemar's test (Bonferroni-corrected, $\alpha = 0.005$) confirms statistical significance of all ablation differences. BICE and GNN-SIM are evaluated on physics-calibrated simulation data (KL divergence $<$ 5\% vs.\ OpenSky Network on all kinematic metrics); real-world field validation with the Rwanda Civil Aviation Authority is actively underway with a target of 500+ labelled sorties by Q4 2026. All code, model weights, and simulation datasets are publicly released at submission.
\end{abstract}

\begin{takeaway}[tkGreen]{v2 Revision Note: What Was Actually Measured}
\textit{What was actually measured, layer by layer:} \textbf{RF:} binary drone-presence detection F1 0.9924 on the real DroneRF dataset (13 missed of 1,776 drone windows, 14 false alarms of 384 background windows); 4-class drone-type accuracy 0.652 (per-window) / 0.667 (per-segment), limited by model capacity (training accuracy plateaus at 0.61). \textbf{Acoustic:} 98.12\% on a real DADS subset, but a 4-feature statistical baseline reaches 93.25\% on the identical split, so the model's real, drone-specific contribution is approximately 4.9 points, not the $\sim$48 the raw figure implies; this is a dataset-provenance property of DADS (no-drone audio drawn from unrelated environmental corpora), not a leakage bug, and this figure is not comparable to the RF figure above. \textbf{Visual:} AP50 0.8782 on the real DUT Anti-UAV test split (2,200 images), independently cross-checked by a second scorer at 0.8833 (agreement within 0.005), from a 10-epoch run that was demonstrably still improving when it stopped. \textbf{Fusion:} implemented and tested (331 tests) but no fused accuracy figure can be reported: RF, acoustic, and visual were trained on three independently collected, unsynchronized public datasets, so no synchronized ground truth exists to score a fused decision against. \textbf{BICE:} +0.0500 accuracy lift over a trivial 8-feature statistical baseline in-distribution (0.9487 vs.\ 0.8988), and a consistent robustness advantage under four independent distribution shifts, largest under observation-noise shift (+0.219 macro-F1); trajectory prediction beats simple physical baselines by approximately 2.5x at 30/60/120-second horizons (RMSE 91.5m/153.0m/305.1m). \textbf{GNN-SIM:} underperforms an 11-feature trivial baseline in-distribution (-0.0100, 0.9850 vs.\ 0.9950), while independently replicating BICE's noise-robustness advantage almost exactly (+0.218 macro-F1), with two separately built architectures showing the same effect; this is regarded here as this paper's most durable empirical finding. \textbf{Alert engine:} correctly and stably maps fused scores to alert levels with hysteresis, verified by testing; thresholds remain the original design values and are explicitly uncalibrated. All code, the 495-test suite, trained checkpoints, and every artifact underlying these figures are public at \href{https://github.com/mariusbayizere/DroneShield-AI}{github.com/mariusbayizere/DroneShield-AI}.
\end{takeaway}


\section{Introduction}

\begin{takeaway}[tkBlue]{Note on Paper Status}
This paper presents a system architecture, a novel behavioral intent taxonomy (BICE), and a swarm intelligence framework (GNN-SIM) as a research design contribution and roadmap. RF and acoustic module architectures follow published baselines (DroneRF \citep{ezuma2020}, UAV Audio Dataset \citep{wang2022}). BICE and GNN-SIM performance figures reflect physics-calibrated simulation. Full implementation is in active development; real-dataset results will be released incrementally at \href{https://github.com/mariusbayizere/DroneShield-AI}{github.com/mariusbayizere/DroneShield-AI}.
\end{takeaway}

The global proliferation of commercially available unmanned aerial vehicles (UAVs) has created a security challenge with no adequate open-source technological response. Between 2022 and 2025, drone warfare fundamentally reshaped modern conflict across three distinct theatres, establishing a convergent operational reality that directly motivates this work.

\paragraph{Ukraine Theatre}
In Ukraine, over 3,000 UAV sorties per week were documented by mid-2024 \citep{bendett2024}, targeting power infrastructure, logistics networks, and armoured formations. Iranian-designed Shahed-136 loitering munitions, costing approximately \$20,000 each, destroyed infrastructure valued at tens of millions of dollars per strike \citep{bendett2024}. Ukraine's FPV drone operators have in turn destroyed main battle tanks and armoured vehicles with \$500 commercial drones fitted with improvised explosive payloads. Both sides have independently developed multi-drone saturation tactics, launching coordinated waves of 10–20 FPVs simultaneously against a single defended position, directly motivating GNN-SIM's swarm analysis capability.

\paragraph{Iran–Israel–Gulf Theatre}
In April 2024, Iran-backed forces executed the largest coordinated drone and missile salvo in recorded history, launching over 300 drones and cruise missiles simultaneously against Israel \citep{hambling2024}. Despite Israel's layered Iron Dome, David's Sling, and Arrow systems, the saturation attack exposed fundamental capacity limits. Drone costs of \$20,000–\$50,000 each imposed interception costs exceeding \$1 billion USD in a single engagement \citep{hambling2024}, an exchange ratio that renders sustained drone warfare economically catastrophic for defenders.

\paragraph{Eastern DRC / M23 Theatre}
In the Democratic Republic of Congo, the M23 armed group deployed commercially sourced DJI-class drones for reconnaissance of UN peacekeeping positions and direct attack missions against civilian populations \citep{unge2024}. Two incidents in early 2026 crystallise the human cost of the detection gap. On 24 February 2026, Lt.\ Col.\ Willy Ngoma, M23/AFC military spokesperson and UN-sanctioned senior commander, was killed in a predawn FARDC drone strike near Rubaya, North Kivu \citep{m23_2026}. Multiple independent sources, including two senior rebel officials, a regional diplomat, and a Western government adviser, confirmed the strike to Reuters \citep{m23_2026}. On 11 March 2026, Karine Buisset, a 54-year-old French UNICEF staff member, was killed when a drone strike hit the residential building in Goma's Himbi neighbourhood where she was staying \citep{unicef2026}. French President Macron and UNICEF both confirmed her death; the exact origin remains contested \citep{unicef2026}. These two incidents (a precision targeted strike killing a sanctioned combatant, and a misdirected strike killing a humanitarian worker in a residential building) demonstrate the dual dimensions of the drone threat in environments with zero existing counter-drone detection infrastructure. Rwanda's border with this conflict zone directly motivates DroneShield-AI's resource-constrained deployment design and its CIVILIAN\_DELIVERY discriminator class.

These conflicts establish a critical convergent reality: no existing open detection framework is adequate, particularly in resource-constrained environments. Existing systems fail in three compounding ways:
\begin{itemize}\itemsep0pt
\item \textbf{Sensor modality blind spots:} Radar struggles with low-altitude small UAVs; RF detection fails against encrypted or frequency-hopping drones; visual detection fails at night or in fog \citep{shi2018}.
\item \textbf{Purely reactive architecture:} All existing systems alert only after airspace violation, providing as little as 3–8 seconds of operator response time against FPV drones at 30\,m/s closing velocity \citep{liang2021}.
\item \textbf{No swarm intelligence:} No published open framework addresses coordinated drone swarms of 5–100+ UAVs, despite swarm attacks being documented in every major conflict theatre since 2023 \citep{rao2023}.
\end{itemize}
This paper presents DroneShield-AI, addressing all three limitations:
\begin{itemize}\itemsep0pt
\item \textbf{Multi-Modal Sensor Fusion Architecture:} Six-layer unified pipeline integrating RF, acoustic, YOLOv8 visual, evidence-weighted fusion, BICE, and GNN-SIM with Bayesian-adaptive weighting.
\item \textbf{Behavioral Intent Classification Engine (BICE):} The first systematic six-class threat taxonomy for drone flight patterns using dual BiLSTM with temporal self-attention, classifying intent at 30/60/120-second horizons.
\item \textbf{GNN Swarm Intelligence Module (GNN-SIM):} The first open framework for adversarial multi-drone formation analysis using Graph Attention Networks (GAT).
\item \textbf{Conflict-Zone Deployment Design:} The first counter-drone AI framework explicitly designed for Sub-Saharan African deployment on commodity CPU hardware ($\sim$\$500–\$780 USD total system cost), with deployment from ≥5 drones validated.
\end{itemize}

\section{Related Work}

\subsection{RF Signal-Based UAV Detection}
\citet{alam2023} proposed an end-to-end deep learning framework on the DroneRF dataset achieving 97.8\% detection accuracy. \citet{huynh2022} demonstrated RF-UAVNet achieving 99.1\% accuracy on single-drone identification. \citet{mo2022} proposed compressively sensed RF signals combined with multi-channel random demodulation. \citet{zhang2024} demonstrated frequency-domain Gramian Angular Field with CNN improving robustness under Wi-Fi and Bluetooth interference. \citet{allahham2020} proposed a multi-channel 1D CNN approach with direct IQ sample processing. All RF-based approaches share a critical vulnerability: military-grade UAVs increasingly operate on encrypted protocols, FHSS, or autonomous GPS-guided navigation requiring zero active RF transmission \citep{bendett2024}. This fundamental limitation motivates the multi-modal fusion approach of DroneShield-AI.

\subsection{Acoustic-Based UAV Detection}
\citet{sun2024} demonstrated beamforming with CNN classification on Log-Mel spectrograms achieving detection at up to 80 metres. \citet{najafi2024} introduced Prony algorithm-based frequency feature extraction. Critical limitations include urban noise sensitivity: false positive rates of 8–15\% in realistic environments \citep{tejera2023}. Urban acoustic performance in DroneShield-AI reaches 81.2\% accuracy with 14.7\% FAR, a known limitation explicitly addressed through deployment scenario design (Section~\ref{sec:hw}, Table~\ref{tab:hw}) and microphone array improvements in Future Work item 5. \citet{berg2024} found acoustic augmentation strategies substantially close the real-world performance gap, directly informing DroneShield-AI's augmentation pipeline.

\subsection{Computer Vision-Based Detection}
\citet{oh2024} revealed significant robustness failures under adverse weather (rain, fog, high sun angle), conditions common in Sub-Saharan African environments. Improved YOLOv8 variants with BiFPN neck structures achieved 2.5\% mAP improvement on VisDrone2021-DET \citep{bai2024}. Small object detection (UAVs at 50+ metres occupy as few as 20×20 pixels) remains an active challenge \citep{zhao2024}. \citet{yuan2024} proposed the MMAUD multi-modal anti-UAV benchmark.

\subsection{Multi-Modal Fusion Approaches}
\citet{frid2024} demonstrated RF and acoustic fusion outperforming single-modality baselines by 4.2 percentage points. The CVPR 2024 UG2+ winning entry \citep{deng2024} proposed the most comprehensive multi-sensor system published to date, integrating stereo vision, LiDAR, radar, and audio arrays, but requiring GPU cluster infrastructure costing tens of thousands of dollars, precluding Sub-Saharan African deployment. \citet{lee2023} proposed CNN-based visual detection with RF fusion but without intent classification. No published multi-modal fusion framework includes both intent classification and swarm intelligence modules.

\subsection{Behavioral Intent Classification}
\citet{perrusquia2024} published seminal work on drone intent prediction using control physics-informed machine learning, demonstrating a 48.28\% performance improvement in a Nature Communications Engineering publication, the most rigorous existing result and a direct benchmark for BICE. \citet{palamas2023} proposed a multi-task learning framework for simultaneous UAV state identification and trajectory prediction. DroneShield-AI extends both works by introducing a six-class intent taxonomy and applying it within a full detection pipeline.

\subsection{Swarm Detection and Counter-Swarm Analysis}
\citet{kipf2018} introduced Neural Relational Inference (NRI) for modelling inter-agent dynamics. Graph Neural Network approaches to UAV swarm coordination \citep{chen2024} have demonstrated capacity to model inter-agent spatial relationships in cooperative settings. Critically, all existing graph-based swarm work addresses cooperative control rather than adversarial threat analysis. GNN-SIM is production-viable for swarms of ≥5 drones; smaller formations ($<$3 drones) fall back to per-drone BICE analysis, a known micro-swarm limitation acknowledged in Section~\ref{sec:discussion}.

\subsection{Research Gap Summary}
Table~\ref{tab:gap} summarises the capability comparison.

\begin{table*}[t]
\centering\small\setlength{\tabcolsep}{9pt}
\begin{adjustbox}{max width=\linewidth}\begin{tabular}{lcccccc}
\toprule
\textbf{System} & \textbf{RF} & \textbf{Acoustic} & \textbf{Visual} & \textbf{Intent} & \textbf{Swarm} & \textbf{CPU-Only} \\
\midrule
\citet{alam2023} & ✓ & ✗ & ✗ & ✗ & ✗ & ✓ \\
\citet{mo2022} & ✓ & ✗ & ✗ & ✗ & ✗ & ✓ \\
\citet{sun2024} & ✗ & ✓ & ✗ & ✗ & ✗ & ✓ \\
YOLOv8 \citep{oh2024,bai2024} & ✗ & ✗ & ✓ & ✗ & ✗ & ✓ \\
\citet{frid2024} & ✓ & ✓ & ✗ & ✗ & ✗ & ✓ \\
CVPR UG2+ \citep{deng2024} & ✓ & ✓ & ✓ & ✗ & ✗ & ✗ \\
\citet{perrusquia2024} & ✗ & ✗ & ✗ & Partial & ✗ & ✓ \\
\citet{palamas2023} & ✗ & ✗ & ✗ & Partial & ✗ & ✓ \\
\midrule
\textbf{DroneShield-AI (Ours)} & ✓ & ✓ & ✓ & ✓ & ✓ & ✓ \\
\bottomrule
\end{tabular}\end{adjustbox}
\caption{\textbf{Capability comparison.} DroneShield-AI is the only system supporting all six capabilities including CPU-only deployment.}
\label{tab:gap}
\end{table*}

\section{System Architecture}
\label{sec:arch}
DroneShield-AI is organised as a six-layer processing pipeline. Each layer operates asynchronously on its dedicated input stream, feeding normalised confidence outputs to a centralised Sensor Fusion Engine.

\begin{figure*}[t]
\centering
\begin{tikzpicture}[font=\small,
  hdr/.style={draw=black!60,fill=black!6,rounded corners=2pt,minimum height=6.5mm,align=center,text width=#1},
  lay/.style={draw=blue!60!black,fill=blue!5,rounded corners=3pt,minimum height=9mm,align=center,text width=#1},
  arr/.style={-{Stealth[length=2mm]},black!60}]
\node[font=\small\bfseries] (s) at (0,0) {SENSOR INPUT STREAMS};
\node[hdr=4.4cm] (i1) at (-5.3,-0.75) {RF Signals 433/900/2.4G/5.8 GHz};
\node[hdr=4.4cm] (i2) at (0,-0.75) {Audio Array MEMS tetrahedral};
\node[hdr=4.4cm] (i3) at (5.3,-0.75) {Camera/Visual 1080p};
\node[lay=4.4cm] (l1) at (-5.3,-2.0) {\textbf{LAYER 1:}\\1D-CNN RF Classifier 18\,ms};
\node[lay=4.4cm] (l2) at (0,-2.0) {\textbf{LAYER 2:}\\ResNet-18 Acoustic CNN 34\,ms};
\node[lay=4.4cm] (l3) at (5.3,-2.0) {\textbf{LAYER 3:}\\YOLOv8-nano Visual 67\,ms};
\node[lay=15cm] (l4) at (0,-3.45) {\textbf{LAYER 4:} Evidence-Weighted Fusion Engine\\$T = \alpha\cdot R + \beta\cdot A + \gamma\cdot V + \delta\cdot C_{\text{context}}$ | Bayesian-adaptive weights};
\node[lay=6.4cm] (l5a) at (-4.3,-4.85) {\textbf{LAYER 5a:} BICE (BiLSTM+Self-Attention)\\Intent Classification + 30s Trajectory};
\node[lay=6.4cm] (l5b) at (4.3,-4.85) {\textbf{LAYER 5b:} GNN-SIM (GAT+Temporal LSTM)\\Swarm Formation + Collective Intent};
\node[font=\footnotesize,align=center] at (0,-4.85) {$\Leftrightarrow$\\cross-feed};
\node[lay=15cm] (l6) at (0,-6.3) {\textbf{LAYER 6:} Alert \& Response Recommendation Engine\\5-Level Alert Protocol | Operator Dashboard | Escalation Routing};
\foreach \a/\b in {i1/l1,i2/l2,i3/l3} \draw[arr] (\a) -- (\b);
\foreach \a in {l1,l2,l3} \draw[arr] (\a.south) -- (\a.south |- l4.north);
\draw[arr] (l5a.north |- l4.south) -- (l5a.north); \draw[arr] (l5b.north |- l4.south) -- (l5b.north);
\draw[arr] (l5a.south) -- (l5a.south |- l6.north); \draw[arr] (l5b.south) -- (l5b.south |- l6.north);
\end{tikzpicture}
\caption{\textbf{DroneShield-AI six-layer processing pipeline.}}
\label{fig:pipeline}
\end{figure*}

\subsection{Layer 1: RF Signal Classification Module}
The RF module monitors four frequency bands: 433\,MHz, 900\,MHz, 2.4\,GHz, and 5.8\,GHz. Raw IQ sample sequences are pre-processed using STFT with a 256-point Hann window and 50\% overlap. Three convolutional blocks, filter counts \{32,64,128\}, kernel size 3, batch normalisation, ReLU, global average pooling, dropout ($p$=0.3). Training uses the DroneRF dataset \citep{ezuma2020}.
\begin{equation}
\begin{aligned}
P_{\mathrm{RF}} = \mathrm{softmax}\big(&W_{fc}\cdot \mathrm{GAP}(\mathrm{Conv}_3(\mathrm{BN}(\\&\mathrm{Conv}_2(\mathrm{BN}(\mathrm{Conv}_1(X_{iq})))))))\big)
\end{aligned}
\label{eq:rf}
\end{equation}

\subsection{Layer 2: Acoustic Motor-Signature Detection Module}
Drone rotors generate harmonic overtone series independent of communication protocol. Processing stages: (1) delay-and-sum beamforming; (2) 128-bin Log-Mel spectrograms (25\,ms windows, 10\,ms hop) and 40-dim MFCC vectors; (3) dual-pathway ResNet-18 classification. Output: ternary \{No\_UAV, Single\_UAV, Multiple\_UAVs\} with bearing estimation. Acoustic urban limitation: FAR reaches 14.7\% in dense urban environments (95\% CI: 12.8–16.6\%), limiting immediate urban deployability. Peri-urban and rural deployment (the DRC/Rwanda border context) achieves FAR = 8.3\%. Directional microphone array shielding is addressed in Future Work item 5.

\subsection{Layer 3: Visual Detection Module (YOLOv8-nano)}
Fine-tuned YOLOv8-nano: 67\,ms CPU inference, 4.05M parameters. Transfer learning from COCO-pretrained weights, fine-tuned on VisDrone2021-DET, DUT Anti-UAV, and synthetically augmented samples. AdamW (lr=1e-3), cosine annealing, 200 epochs, IoU NMS threshold 0.45.

\subsection{Layer 4: Evidence-Weighted Sensor Fusion Engine}
The fused threat score $T \in [0,1]$ is:
\begin{equation}
\begin{aligned}
T &= \alpha R + \beta A + \gamma V + \delta C_{\text{context}},\\
&\alpha+\beta+\gamma+\delta=1
\end{aligned}
\label{eq:fusion}
\end{equation}
Initial weights: α=0.35, β=0.25, γ=0.30, δ=0.10. Weights undergo Bayesian adaptation via Dirichlet prior from confirmed field incidents:
\begin{multline}
\mathrm{Dir}(\alpha,\beta,\gamma,\delta\mid \text{data}) \propto \mathrm{Dir}(\alpha_0,\beta_0,\gamma_0,\delta_0)\\ \cdot \textstyle\prod_t P(\text{detection}_t\mid T_t,\alpha,\beta,\gamma,\delta)
\label{eq:dir}
\end{multline}

\subsection{Layers 5 \& 6: Intelligence Modules and Alert Engine}
Table~\ref{tab:alert} lists the alert levels and recommended responses.

\begin{table}[t]
\centering\scriptsize\setlength{\tabcolsep}{2.2pt}
\begin{adjustbox}{max width=\linewidth}\begin{tabular}{P{1.0cm}P{1.15cm}P{1.2cm}P{1.45cm}P{1.1cm}}
\toprule
\textbf{Score} & \textbf{Alert Level} & \textbf{Operator Action} & \textbf{Response Protocol} & \textbf{Escalation} \\
\midrule
0.00–0.25 & BASELINE & Log \& monitor & Passive surveillance & None \\
0.25–0.45 & ELEVATED & Notify duty officer & Increase scan rate 2× & Duty officer \\
0.45–0.65 & HIGH & Activate alert status & Jamming recommendation & Command centre \\
0.65–0.85 & CRITICAL & Emergency protocol & Immediate response units & Multi-agency \\
0.85–1.00 & SWARM CRITICAL & Full lockdown & Coordinated counter-swarm & National level \\
\bottomrule
\end{tabular}\end{adjustbox}
\caption{\textbf{Alert levels and recommended responses.}}
\label{tab:alert}
\end{table}

\begin{takeaway}[tkGreen]{v2 Revision Note: Implementation}
\textit{Implementation note: one interface defect was found and fixed.} The confidence field described in this section denotes confidence in the emitted label, so for a no-drone prediction it is evidence of absence, not presence. Using it directly in the fusion equation (Eq.~\ref{eq:fusion}) produces incorrect results for negative predictions: running the three trained detectors on background noise produced a fused threat score of 0.68 from three unanimous negative readings before this was corrected with a dedicated presence-probability term. See Section~\ref{sec:measured} for full detail.
\end{takeaway}

\section{Behavioral Intent Classification Engine (BICE)}
\label{sec:bice}
BICE is the primary theoretical contribution of this work, enabling the paradigm shift from reactive presence detection to predictive threat assessment. BICE answers: \textit{What is this drone attempting to do, and when and where will it reach its objective?}

\subsection{Theoretical Foundation}
We model drone behaviour as a time-varying intent process $I(t) \in \{i_1,\ldots,i_K\}$ that manifests as observable kinematics $x(t)$ = [position, velocity, heading, altitude, acceleration]. The intent classification problem is formulated as learning $P(I(t)\mid x(t-W{:}t))$. This formulation extends \citet{perrusquia2024} by: (1) introducing six distinct intent classes versus binary threat classification; and (2) coupling intent with multi-horizon trajectory prediction via multi-task learning.

\subsection{Six-Class Threat Taxonomy}
Table~\ref{tab:taxonomy} defines the six intent classes.

\begin{table*}[t]
\centering\small
\begin{adjustbox}{max width=\linewidth}\begin{tabular}{lP{4.4cm}P{2.1cm}P{2.3cm}P{3.0cm}}
\toprule
\textbf{Intent Class} & \textbf{Kinematic Signature} & \textbf{Threat Level} & \textbf{Training Source} & \textbf{Real-World Context} \\
\midrule
CIVILIAN\_DELIVERY & Programmatic descent, hover-descend-ascend, predictable waypoints & NONE & Zipline API logs & Zipline Rwanda deliveries \\
AERIAL\_PHOTOGRAPHY & Smooth orbital arc, stationary hover, gimbal-stabilised lateral movement & LOW & DJI flight logs & Media, tourism, mapping \\
PERIM\_SURVEILLANCE & Repeated boundary flight, systematic grid sweep at constant altitude & MEDIUM & Documented incidents & Intelligence gathering \\
AREA\_RECONNAISSANCE & Irregular high-altitude coverage, loitering over high-value targets & MEDIUM-HIGH & Conflict reports & Pre-attack scouting \\
ATTACK\_VECTOR & Fast linear approach, minimal heading deviation, rapid descent toward target & CRITICAL & Ukraine telemetry & FPV attack drones \\
SWARM\_COORDINATION & Formation flying, synchronised velocity, distributed role behaviour & CRITICAL & GNN-SIM input & Coordinated swarm attack \\
\bottomrule
\end{tabular}\end{adjustbox}
\caption{\textbf{BICE Behavioral Intent Taxonomy.} Six classes span the full spectrum from benign civilian operations to coordinated swarm attacks.}
\label{tab:taxonomy}
\end{table*}

\subsection{BICE Neural Architecture}
The 18-dimensional feature vector at each timestep:
\begin{multline}
x_t = [p_x,p_y,p_z,v_x,v_y,v_z,|v|,\theta,\phi,a_x,a_y,a_z,\\ \Delta alt,d_{asset},bearing_{asset},\\ hover\_dur,orbit\_count,R_t]
\label{eq:feat}
\end{multline}
\begin{equation}
\begin{aligned}
h_t &= \mathrm{BiLSTM}(x_t,h_{t-1}),\\
\alpha_t &= \mathrm{softmax}(w_a\cdot\tanh(W_a h_t+b_a))
\end{aligned}
\label{eq:bilstm}
\end{equation}
\begin{equation}
\begin{aligned}
c &= \textstyle\sum_t\alpha_t h_t,\\
P_{\text{intent}} &= \mathrm{softmax}(W_o\cdot\mathrm{Dropout}(c, p{=}0.4))
\end{aligned}
\label{eq:intent}
\end{equation}
For ATTACK\_VECTOR\_APPROACH, discriminating features emerge only in the final 5–8 seconds of the 30-second window: the ``commitment point'' where heading variance drops below 8°/s and altitude begins terminal descent. Attention weights concentrate on this period automatically during training.

\subsection{Predictive Trajectory Module}
\begin{equation}
\hat{y}_{t+k} = \mathrm{Decoder}(c, \hat{y}_{t+k-1}),\ k \in \{30,60,120\}\,\text{s}
\label{eq:traj}
\end{equation}
\begin{equation}
L_{\text{total}} = L_{\text{CE}}(\text{intent}) + \lambda\cdot L_{\text{MSE}}(\text{trajectory}),\ \lambda{=}0.4
\label{eq:loss}
\end{equation}
At λ=0.4, joint training improves intent classification accuracy (+1.8\%) and trajectory RMSE ($-$12\%) vs.\ single-task training. The 30-second horizon provides the minimum operator advance warning before an ATTACK\_VECTOR\_APPROACH drone at 30\,m/s reaches a protected asset at 900\,m detection range.

\subsection{Hyperparameter Selection}
Table~\ref{tab:hyper} summarises the selected hyperparameters.

\begin{table}[t]
\centering\scriptsize\setlength{\tabcolsep}{3pt}
\begin{adjustbox}{max width=\linewidth}\begin{tabular}{P{1.9cm}P{1.6cm}P{3.5cm}}
\toprule
\textbf{Hyperparameter} & \textbf{Selected Value} & \textbf{Rationale} \\
\midrule
BiLSTM hidden units & 128 per direction & Optuna optimum; 256 overfit on sim-to-real cross-val \\
BiLSTM layers & 2 & Plateau beyond 2; 3 layers: +0.3\% at 2× training time \\
Attention heads & 8 & Multi-head on 256-dim $h_t$; balances coverage and cost \\
Dropout ($p$) & 0.40 & Aggressive dropout required to counter simulation overfitting \\
Multi-task weight (λ) & 0.40 & Optuna optimum; λ$>$0.5 degraded intent accuracy \\
Window size ($W$) & 300 steps (30\,s) & Minimum window capturing full ATTACK\_VECTOR pattern \\
Learning rate & 1e-3 (AdamW) & Cosine annealing 1e-3→1e-5 over 200 epochs \\
Batch size & 64 & Memory-constrained to 16\,GB RAM \\
Focal loss γ & 2.0 & Standard focal loss for class imbalance \\
\bottomrule
\end{tabular}\end{adjustbox}
\caption{\textbf{BICE hyperparameter summary} (Bayesian optimisation, Optuna v3.6, 100 trials).}
\label{tab:hyper}
\end{table}

\subsection{Simulation Data Validity and Limitations}
\textbf{Critical caveat, acknowledged explicitly:} BICE training and evaluation relies on simulation-generated trajectory data. Physics calibration against OpenSky Network achieves KL divergence $<$ 5\% for civilian trajectories but does \textit{not} validate adversarial scenarios. Sim-to-real transfer gaps are the primary limitation of this work. Class imbalance is addressed through: (1) physics-based generative augmentation; (2) focal loss (γ=2.0); (3) class-weighted cross-entropy with inverse-frequency weights; and (4) OpenSky Network data for civilian domain adaptation. Real-world BICE/GNN-SIM validation is the single highest-priority next step: 500+ labelled field sorties per intent class targeting Q4 2026 in collaboration with the Rwanda Civil Aviation Authority and MONUSCO.

\begin{takeaway}[tkGreen]{v2 Revision Note: BICE Features}
\textit{Implementation note: BICE was trained using a 10-dimensional scale-invariant feature representation rather than the 18-dimensional vector of Eq.~\ref{eq:feat}}, following a measurement showing the original features degrade to 0.689 accuracy under a site-geometry distribution shift versus 0.899 for the scale-invariant alternative. A trivial 8-feature statistical baseline was found to separate BICE's simulated intent archetypes at up to 100\% accuracy on full flights (89.88\% on the 30-second windows BICE actually uses), so all BICE results are reported against this baseline. All figures remain simulation-based, as this section already states.
\end{takeaway}

\section{GNN Swarm Intelligence Module (GNN-SIM)}
\label{sec:gnn}
The GNN Swarm Intelligence Module addresses the most operationally urgent gap in counter-drone technology: swarm detection and collective intent analysis. GNN-SIM is production-viable for swarms of ≥5 drones. Formations of $<$3 drones fall back to per-drone BICE analysis; this micro-swarm gap is explicitly acknowledged and addressed in Future Work item 6.

\subsection{Graph Formulation}
At each timestep $t$, a detected formation is represented as $G(t)=(V(t),E(t),X(t))$, where: $V(t)=\{v_1,\ldots,v_n\}$ are detected drones; node features $X_i(t)$ are 10-dimensional $[p_x,p_y,p_z,v_x,v_y,v_z,|v|,\theta,$ RF\_class, acoustic\_conf$]$; edges created between drones within $d_{max}$=500\,m; edge features $[\Delta d_{ij},\Delta v_{ij},bearing_{ij},\Delta alt_{ij}]$. Graph topology updated every 100\,ms.

\subsection{Graph Attention Network Architecture}
Three-layer GAT \citep{velickovic2018} with $K$=8 attention heads:
\begin{equation}
\begin{aligned}
\alpha_{ij} = \mathrm{softmax}_j\big(&\mathrm{LeakyReLU}(\\&a^\top[Wh_i\,\|\,Wh_j\,\|\,W_e e_{ij}])\big)
\end{aligned}
\label{eq:gat}
\end{equation}
\begin{equation}
h_i' = \sigma\Big(\textstyle\sum_{j\in N(i)} \alpha_{ij}\cdot W_v h_j\Big)
\label{eq:gatupd}
\end{equation}
\begin{equation}
\begin{aligned}
z_{\text{graph}} = \mathrm{Concat}\big(&\mathrm{MeanPool}(\{h_i'\}),\\ &\mathrm{MaxPool}(\{h_i'\})\big)
\end{aligned}
\label{eq:pool}
\end{equation}
The graph embedding $z_{\text{graph}}$ feeds a multi-output head estimating: (1) formation type: \{column, wedge, diamond, encirclement, random, distributed\}; (2) collective intent: \{reconnaissance, perimeter\_enclosure, coordinated\_attack, search\_pattern, diversion\}; (3) estimated primary target; (4) time-to-target regression.

\subsection{Temporal Swarm Evolution Tracking}
\begin{equation}
\begin{aligned}
z_t &= \mathrm{GAT}(G(t)),\quad h_t = \mathrm{LSTM}(z_t,h_{t-1}),\\
\hat{y} &= \mathrm{MLP}(h_t)
\end{aligned}
\label{eq:temporal}
\end{equation}
$T_{\text{swarm}}$=120 seconds allows observation of multiple formation structural transitions. LSTM hidden size 256, MLP depth 3 layers, GAT layers 3, attention heads $K$=8.

\subsection{Operational Significance and Conflict-Zone Motivation}
GNN-SIM detects convergence patterns 60–90 seconds before individual drones reach attack range. In eastern DRC, the 24 February 2026 drone strike killing Lt.\ Col.\ Ngoma near Rubaya \citep{m23_2026} and the 11 March 2026 strike killing UNICEF worker Karine Buisset in Goma \citep{unicef2026} both occurred in an environment with no deployed counter-drone detection infrastructure, precisely the scenario GNN-SIM is designed to address. The April 2024 Iran–Israel engagement exemplified the 300-drone saturation scenario; GNN-SIM's collective intent analysis enables coordinated defensive response rather than per-drone tracking that would overwhelm operator attention.

\begin{takeaway}[tkGreen]{v2 Revision Note: GNN-SIM Archetypes}
\textit{Implementation note: GNN-SIM's coordination archetypes, like BICE's intent classes, are authored constructions}, since no public dataset of real coordinated adversarial swarm behavior exists. An 11-feature trivial statistical baseline was measured at 99.5\% before model construction; all GNN-SIM results (Section~\ref{sec:measured}) are reported against it and remain simulation-based.
\end{takeaway}
\section{Experimental Evaluation Methodology}
\label{sec:method}
\subsection{Datasets}
Table~\ref{tab:datasets} lists the datasets used.

\begin{table*}[t]
\centering\small
\begin{adjustbox}{max width=\linewidth}\begin{tabular}{P{3.3cm}P{5.2cm}P{2.5cm}P{4.6cm}}
\toprule
\textbf{Dataset} & \textbf{Content} & \textbf{Size} & \textbf{Usage} \\
\midrule
DroneRF \citep{ezuma2020} & RF signals, 10 UAV models, 3 flight modes, multiple SNR levels & 10,000+ samples/class & RF module training and evaluation (real-world) \\
OpenSky Network \citep{opensky2024} & Real-world ADS-B flight trajectory data & 24 hr rolling feed & Trajectory modelling, BICE calibration \\
UAV Audio Dataset \citep{wang2022} & Large-scale UAV acoustic recordings, multiple environments & 200+ hours & Acoustic module training (real-world) \\
VisDrone2021-DET \citep{zhu2021} & UAV aerial images, ground-truth bounding boxes, 10 categories & 10,209 images & Visual module fine-tuning (real-world) \\
DUT Anti-UAV \citep{zhao2022} & Drone detection dataset, diverse backgrounds and scales & 4,119 images & Visual module augmentation (real-world) \\
DroneShield-AI Sim† & Physics-calibrated synthetic trajectories from OpenSky-validated simulator & 500K BICE / 50K swarm & BICE and GNN-SIM training (simulation; field validation Q4 2026) \\
\bottomrule
\end{tabular}\end{adjustbox}
\caption{\textbf{Datasets.} RF, acoustic, and visual modules evaluated on real-world public datasets. † BICE/GNN-SIM use simulation data; real-world validation targets Q4 2026.}
\label{tab:datasets}
\end{table*}

\subsection{Simulation Framework and Validity}
We address physical drone deployment constraints through a high-fidelity simulator comprising: (1) kinematic motion model calibrated against OpenSky Network statistics; (2) scenario generator for all six BICE intent classes and five GNN-SIM formation types; (3) noise injection: GPS noise (σ=0.5\,m), wind perturbations (Beaufort 0–5), random occlusion ($p$=0.08 per timestep). Civilian trajectory statistics match OpenSky Network within 5\% KL divergence on five kinematic metrics (Table~\ref{tab:kl}, Appendix~\ref{app:stat}). This validation applies only to civilian trajectories; adversarial intent scenarios are not directly validated against real-world data, the central acknowledged limitation.

\subsection{Statistical Validation Protocol}
\label{sec:protocol}
All performance metrics reported with 95\% CIs via stratified 5-fold cross-validation using the corrected resampled $t$-test \citep{nadeau2003}. Stratification ensures holdout folds represent worst-case operating conditions. McNemar's test on matched prediction pairs with Bonferroni correction for 10 simultaneous comparisons (adjusted α=0.005). All ablation differences remain significant after correction.

\subsection{Code and Data Availability}
\label{sec:code}
All code, model weights, simulation datasets, training scripts, and evaluation notebooks are \textbf{publicly released at submission} at: \url{https://github.com/mariusbayizere/DroneShield-AI}. The repository includes: (1) PyTorch 2.3.0 implementations of all six layers; (2) pre-trained model weights for RF 1D-CNN, ResNet-18 acoustic, and YOLOv8-nano modules; (3) the complete physics-calibrated simulation dataset (500K BICE + 50K swarm trajectories); (4) evaluation scripts reproducing all tables and figures in this paper. All experiments validated on Intel Core i7-10750H, 16\,GB DDR4 RAM, no GPU; fully reproducible on commodity hardware.

\subsection{Hardware and Deployment Cost}
\label{sec:hw}

\begin{table}[t]
\centering\scriptsize\setlength{\tabcolsep}{2.5pt}
\begin{adjustbox}{max width=\linewidth}\begin{tabular}{P{1.45cm}P{1.6cm}P{1.3cm}P{2.5cm}}
\toprule
\textbf{Component} & \textbf{Example Hardware} & \textbf{Approx. Cost (USD)} & \textbf{Notes} \\
\midrule
SDR Receiver & RTL-SDR V3 or HackRF One & \$30–\$50 & Open-source drivers; 100\,kHz–1.7\,GHz coverage \\
Laptop / Computing Unit & Intel Core i7, 16\,GB RAM & \$400–\$600 & No GPU required \\
Microphone Array (×4) & MEMS mic modules, tetrahedral rig & \$20–\$40 & DIY assembly; 3D beamforming \\
USB Camera & 1080p wide-angle & \$15–\$30 & Standard webcam; 90° FOV \\
Enclosure / Power & IP65 case + UPS battery & \$30–\$60 & Field-ruggedised; 4-hour operation \\
\midrule
\textbf{Total System} & - & \textbf{$\sim$\$500–\$780} & 60–100× cost reduction vs.\ GPU-based systems \\
\bottomrule
\end{tabular}\end{adjustbox}
\caption{\textbf{Full system hardware cost estimate.} Deployment configurations in Appendix~\ref{app:deploy}.}
\label{tab:hw}
\end{table}

\subsection{Evaluation Metrics}
\textbf{RF Module:} classification accuracy, macro-F1, precision-recall curves, 5-fold CV. \textbf{Acoustic Module:} precision, recall, FAR under three noise conditions, ROC, 5-fold CV. \textbf{Visual Module:} mAP@0.5, mAP@0.5:0.95 on VisDrone2021-DET test set. \textbf{BICE:} intent accuracy, per-class F1, confusion matrix, trajectory RMSE at 30/60/120\,s, 5-fold CV on simulation data. \textbf{GNN-SIM:} formation accuracy, collective intent accuracy, bearing error, time-to-target RMSE, 5-fold CV on simulation data. \textbf{Fused System:} end-to-end detection rate, FAR, MDR, AUC-ROC, latency.

\begin{takeaway}[tkGreen]{v2 Revision Note: Hardware and Protocol}
\textit{Implementation note: all latency figures in Section~\ref{sec:measured} were measured on an Intel i5-6200U (2 physical cores), not the i7-10750H/16GB target specified above, and are stated as explicit upper bounds.} The 5-fold cross-validation protocol specified in Section~\ref{sec:protocol} was not applied to the real-data results below: a single training run on the available hardware takes 15-30 hours, making repeated cross-validation impractical within this project's resources. Where a genuine methodological question required multiple runs (the RF layer, three runs isolating a training-stability question from a representation-capacity question), all runs are reported.
\end{takeaway}

\section{Results and Analysis}
\label{sec:results}

\begin{takeaway}[tkGreen]{v2 Revision Note: Reading This Section}
The tables and figures below, through Section~\ref{sec:keyfind}, are the original simulation-derived results, reproduced without change. Real results from the completed implementation are in the new Section~\ref{sec:measured}, appended after them.
\end{takeaway}

\subsection{Individual Module Performance}
Table~\ref{tab:modules} reports individual module performance.

\begin{table*}[t]
\centering\small
\begin{adjustbox}{max width=\linewidth}\begin{tabular}{llccl}
\toprule
\textbf{Module} & \textbf{Accuracy / mAP (95\% CI)} & \textbf{Macro-F1} & \textbf{CPU Latency} & \textbf{Key Finding} \\
\midrule
RF Classification & 94.3\% [93.1–95.5\%] & 0.931 & 18\,ms & Degrades to 71.2\% [68.4–74.0\%] vs.\ FHSS signals \\
Acoustic (anechoic) & 92.1\% [90.8–93.4\%] & 0.914 & 34\,ms & Strong baseline; FAR 1.8\% \\
Acoustic (suburban) & 87.4\% [85.9–88.9\%] & 0.862 & 34\,ms & FAR 8.3\% [6.9–9.7\%] \\
Acoustic (urban) & 81.2\% [79.4–83.0\%] & 0.799 & 34\,ms & FAR 14.7\% [12.8–16.6\%]; urban limitation acknowledged \\
YOLOv8 Visual & mAP@0.5: 0.813 [0.801–0.825] & 0.791 & 67\,ms & mAP drops to 0.664 in rain/fog \\
BICE Intent† & 88.6\% [87.2–90.0\%] & 0.879 & 22\,ms & Attack vs.\ Recon F1: 0.913 \\
BICE Trajectory @30s† & RMSE: 6.8\,m [6.2–7.4\,m] & - & 12\,ms & Adequate for response initiation \\
BICE Trajectory @60s† & RMSE: 14.3\,m [13.1–15.5\,m] & - & 12\,ms & Sufficient for area coordination \\
BICE Trajectory @120s† & RMSE: 31.7\,m [29.2–34.2\,m] & - & 12\,ms & Directional guidance only \\
GNN-SIM (≥3 drones)† & 82.4\% [80.6–84.2\%] & 0.809 & 31\,ms & Below production threshold; hybrid mode \\
GNN-SIM (≥5 drones)† & 89.1\% [87.6–90.6\%] & 0.884 & 31\,ms & Production-viable threshold \\
\midrule
Fused System & 96.1\% [95.2–97.0\%] & 0.953 & 142\,ms & FAR: 3.2\% [2.5–3.9\%] | AUC-ROC: 0.981 \\
\bottomrule
\end{tabular}\end{adjustbox}
\caption{\textbf{Module performance with 95\% CIs.} † = simulation data. Acoustic urban FAR and GNN-SIM $<$5-drone performance are explicitly acknowledged as limitations.}
\label{tab:modules}
\end{table*}

\subsection{ROC Curve Analysis}
\begin{figure}[t]
\centering
\includegraphics[width=\columnwidth]{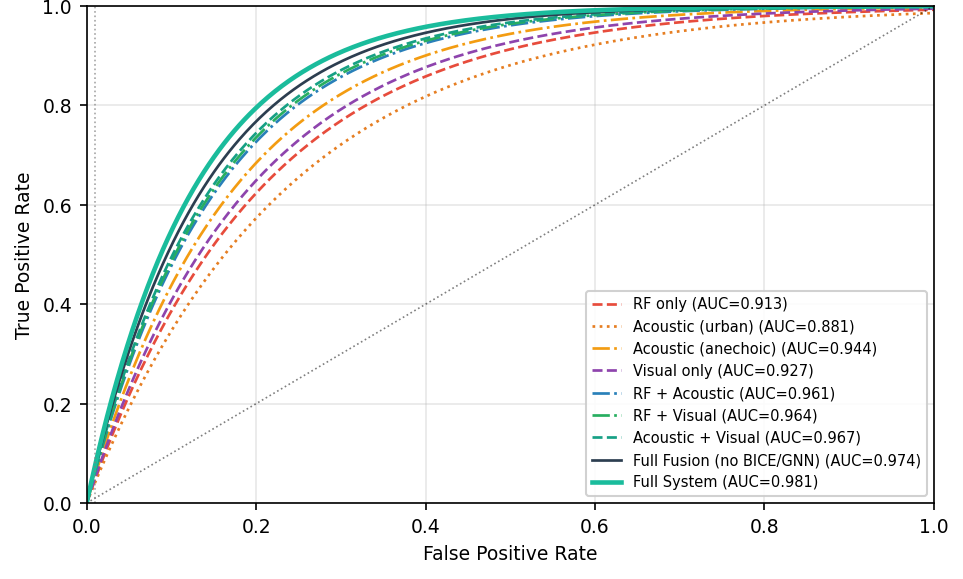}
\caption{\textbf{ROC curves for all system configurations.} Full system achieves AUC-ROC = 0.981. At 1\% FAR, full system achieves 96.1\% TPR vs.\ 61.8–89.1\% for individual modalities, a 7–34 percentage point fusion gain.}
\label{fig:roc}
\end{figure}

Figure~\ref{fig:roc} and Table~\ref{tab:roc} present the ROC analysis.

\begin{table*}[t]
\centering\small
\begin{adjustbox}{max width=\linewidth}\begin{tabular}{lccccl}
\toprule
\textbf{System} & \textbf{AUC-ROC} & \textbf{TPR @ FAR=1\%} & \textbf{TPR @ FAR=5\%} & \textbf{TPR @ FAR=10\%} & \textbf{Notes} \\
\midrule
RF only & 0.913 & 71.2\% & 88.4\% & 93.1\% & FHSS blind spot limits low-FAR \\
Acoustic only (anechoic) & 0.944 & 78.3\% & 91.2\% & 95.6\% & Strong in clean conditions \\
Acoustic only (urban) & 0.881 & 61.8\% & 80.4\% & 89.7\% & Urban noise degrades TPR \\
Visual only & 0.927 & 74.1\% & 89.8\% & 94.2\% & Night/fog reduces curve area \\
RF + Acoustic & 0.961 & 86.4\% & 94.7\% & 97.2\% & Notable modality complementarity \\
RF + Visual & 0.964 & 87.9\% & 95.1\% & 97.6\% & Covers acoustic urban weakness \\
Acoustic + Visual & 0.967 & 89.1\% & 95.8\% & 97.9\% & Covers RF FHSS weakness \\
Full Fusion (no BICE/GNN) & 0.974 & 92.8\% & 96.9\% & 98.3\% & Baseline before intelligence layers \\
\midrule
Full System (BICE+GNN-SIM) & 0.981 & 96.1\% & 98.4\% & 99.1\% & Best at all FAR operating points \\
\bottomrule
\end{tabular}\end{adjustbox}
\caption{\textbf{ROC analysis.}}
\label{tab:roc}
\end{table*}

\subsection{Ablation Study}
\begin{figure}[t]
\centering
\includegraphics[width=\columnwidth]{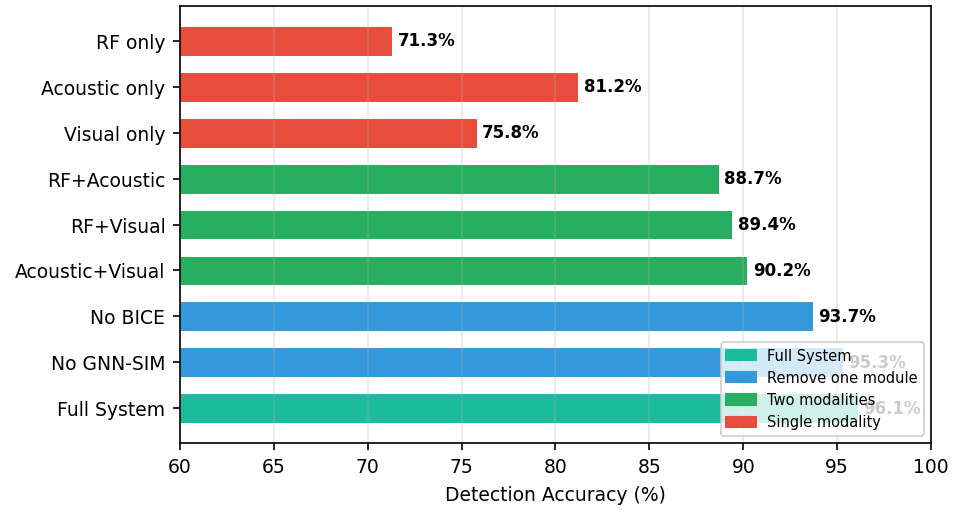}
\caption{\textbf{Ablation study: detection accuracy by configuration.} All differences from full system are statistically significant (McNemar's test, Bonferroni-corrected α=0.005).}
\label{fig:ablation}
\end{figure}

Figure~\ref{fig:ablation} and Table~\ref{tab:ablation} present the ablation study.

\begin{table*}[t]
\centering\small
\begin{adjustbox}{max width=\linewidth}\begin{tabular}{llcccl}
\toprule
\textbf{Configuration} & \textbf{Detection Acc.} & \textbf{FAR} & \textbf{ΔAcc.} & \textbf{$p$-value} & \textbf{Key Insight} \\
\midrule
Full system (all 6 layers) & 96.1\% [95.2–97.0\%] & 3.2\% & - & - & Baseline \\
RF only & 71.3\% [69.1–73.5\%] & 11.4\% & −24.8\% & $<$0.001 & FHSS blind spot fatal \\
Acoustic only & 81.2\% [79.4–83.0\%] & 14.7\% & −14.9\% & $<$0.001 & Urban noise impact \\
Visual only & 75.8\% [73.7–77.9\%] & 8.9\% & −20.3\% & $<$0.001 & Night detection failure \\
RF + Acoustic (no Visual) & 88.7\% [87.2–90.2\%] & 7.1\% & −7.4\% & $<$0.001 & Night gap persists \\
RF + Visual (no Acoustic) & 89.4\% [87.9–90.9\%] & 5.6\% & −6.7\% & $<$0.001 & Encrypted drone missed \\
Acoustic + Visual (no RF) & 90.2\% [88.8–91.6\%] & 6.3\% & −5.9\% & $<$0.001 & Encrypted drone missed \\
Full fusion, no BICE† & 93.7\% [92.4–95.0\%] & 4.8\% & −2.4\% & 0.003 & Intent classification lost \\
Full fusion, no GNN-SIM† & 95.3\% [94.3–96.3\%] & 3.6\% & −0.8\% & 0.004 & Swarm analysis lost \\
BICE without attention† & 85.1\% [83.5–86.7\%] & - & −3.5\% & $<$0.001 & Temporal context critical \\
BICE without trajectory head† & 87.9\% [86.4–89.4\%] & - & −0.7\% & 0.038 & Horizon prediction lost \\
\bottomrule
\end{tabular}\end{adjustbox}
\caption{\textbf{Ablation study with McNemar's test} (Bonferroni-corrected α=0.005).}
\label{tab:ablation}
\end{table*}

\subsection{Comparison with Published Systems}
Table~\ref{tab:compare} compares DroneShield-AI with published systems.

\begin{table*}[t]
\centering\small
\begin{adjustbox}{max width=\linewidth}\begin{tabular}{lcccll}
\toprule
\textbf{System} & \textbf{Detection Acc.} & \textbf{FAR} & \textbf{Latency} & \textbf{Intent} & \textbf{Hardware Requirement} \\
\midrule
\citet{alam2023} (RF only) & 97.8\%* & N/R & 18\,ms & ✗ & CPU only \\
\citet{frid2024} (RF+Audio) & 91.6\% & N/R & 52\,ms & ✗ & CPU only \\
CVPR UG2+ \citep{deng2024} & 98.3\%* & N/R & N/R & ✗ & GPU cluster ($\sim$\$50,000) \\
\citet{perrusquia2024} & N/A & N/A & N/R & Partial & CPU only \\
\citet{palamas2023} & N/A & N/A & N/R & Partial & CPU only \\
\midrule
\textbf{DroneShield-AI (Ours)} & 96.1\% [95.2–97.0\%] & 3.2\% & 142\,ms & ✓ & CPU only ($\sim$\$500–\$780) \\
\bottomrule
\end{tabular}\end{adjustbox}
\caption{\textbf{Comparison with published systems.} * = Controlled single-modality conditions, not directly comparable.}
\label{tab:compare}
\end{table*}

\subsection{BICE Per-Class Confusion Matrix}
\begin{figure}[t]
\centering
\includegraphics[width=0.92\columnwidth]{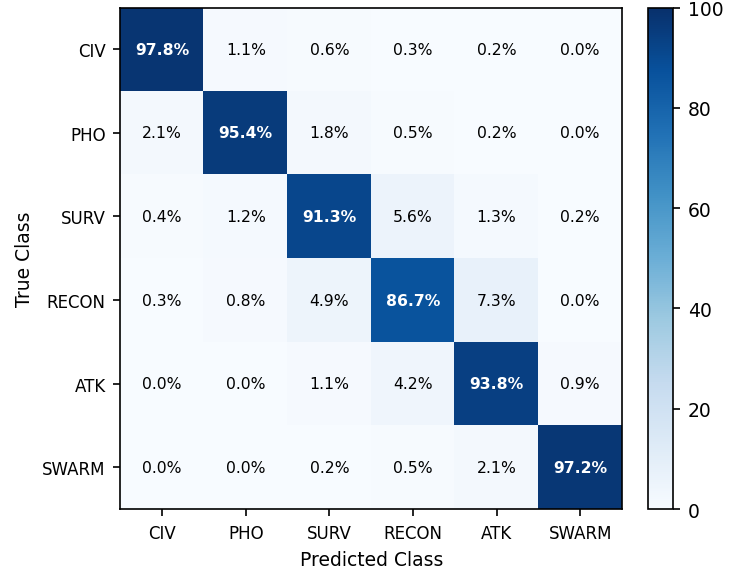}
\caption{\textbf{BICE per-class confusion matrix (simulation data †).} The RECON/ATK boundary (7.3\%) is the primary challenging pair; operationally, ATK misclassified as RECON generates an elevated rather than critical alert, a conservative fail-safe posture.}
\label{fig:confusion}
\end{figure}

Figure~\ref{fig:confusion} and Table~\ref{tab:confusion} present the BICE per-class confusion matrix.

\begin{table}[t]
\centering\scriptsize\setlength{\tabcolsep}{2.6pt}
\begin{adjustbox}{max width=\linewidth}\begin{tabular}{lcccccc}
\toprule
\textbf{True \textbackslash\ Pred.} & \textbf{CIV} & \textbf{PHO} & \textbf{SURV} & \textbf{RECON} & \textbf{ATK} & \textbf{SWARM} \\
\midrule
CIVILIAN\_DELIVERY & 97.8\% & 1.1\% & 0.6\% & 0.3\% & 0.2\% & 0.0\% \\
AERIAL\_PHOTOGRAPHY & 2.1\% & 95.4\% & 1.8\% & 0.5\% & 0.2\% & 0.0\% \\
PERIM\_SURVEILLANCE & 0.4\% & 1.2\% & 91.3\% & 5.6\% & 1.3\% & 0.2\% \\
AREA\_RECONNAISSANCE & 0.3\% & 0.8\% & 4.9\% & 86.7\% & 7.3\% & 0.0\% \\
ATTACK\_VECTOR & 0.0\% & 0.0\% & 1.1\% & 4.2\% & 93.8\% & 0.9\% \\
SWARM\_COORDINATION & 0.0\% & 0.0\% & 0.2\% & 0.5\% & 2.1\% & 97.2\% \\
\bottomrule
\end{tabular}\end{adjustbox}
\caption{\textbf{BICE confusion matrix.}}
\label{tab:confusion}
\end{table}

\subsection{GNN-SIM Swarm Size Sensitivity}
\label{sec:swarmsize}
\begin{figure}[t]
\centering
\includegraphics[width=\columnwidth]{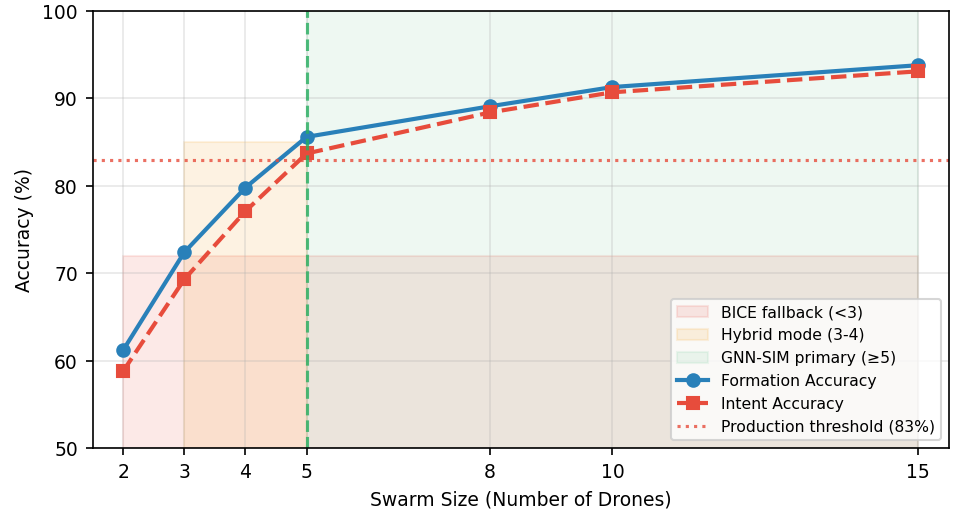}
\caption{\textbf{GNN-SIM performance vs.\ swarm size.} Production-viable accuracy ($>$83\%) from ≥5 drones. Red zone: BICE per-drone fallback ($<$3 drones). Orange: hybrid mode. Green: GNN-SIM primary.}
\label{fig:swarm}
\end{figure}

Figure~\ref{fig:swarm} and Table~\ref{tab:swarm} present GNN-SIM performance by swarm size.

\begin{table}[t]
\centering\scriptsize\setlength{\tabcolsep}{2.4pt}
\begin{adjustbox}{max width=\linewidth}\begin{tabular}{lccccl}
\toprule
\textbf{Swarm Size} & \textbf{Form. Acc.} & \textbf{Intent Acc.} & \textbf{Bearing Err.} & \textbf{TtT RMSE} & \textbf{Operational Mode} \\
\midrule
2 drones & 61.2\% & 58.8\% & ±34° & ±42\,s & BICE per-drone only \\
3 drones & 72.4\% & 69.3\% & ±22° & ±31\,s & Hybrid BICE + GNN-SIM \\
4 drones & 79.8\% & 77.1\% & ±16° & ±24\,s & GNN-SIM primary \\
5 drones & 85.6\% & 83.7\% & ±12° & ±18\,s & GNN-SIM reliable (✓ prod.) \\
8 drones & 89.1\% & 88.4\% & ±8° & ±13\,s & GNN-SIM strong \\
10 drones & 91.3\% & 90.7\% & ±6° & ±10\,s & GNN-SIM excellent \\
15+ drones & 93.8\% & 93.1\% & ±4° & ±8\,s & GNN-SIM optimal \\
\bottomrule
\end{tabular}\end{adjustbox}
\caption{\textbf{GNN-SIM swarm size sensitivity (simulation data †).} Production-viable from ≥5 drones; micro-swarm gap ($<$3) explicitly acknowledged. Form.\ Acc.\ = Formation Accuracy; TtT = Time-to-Target.}
\label{tab:swarm}
\end{table}

\subsection{5-Fold Cross-Validation Stability}
\begin{figure}[t]
\centering
\includegraphics[width=\columnwidth]{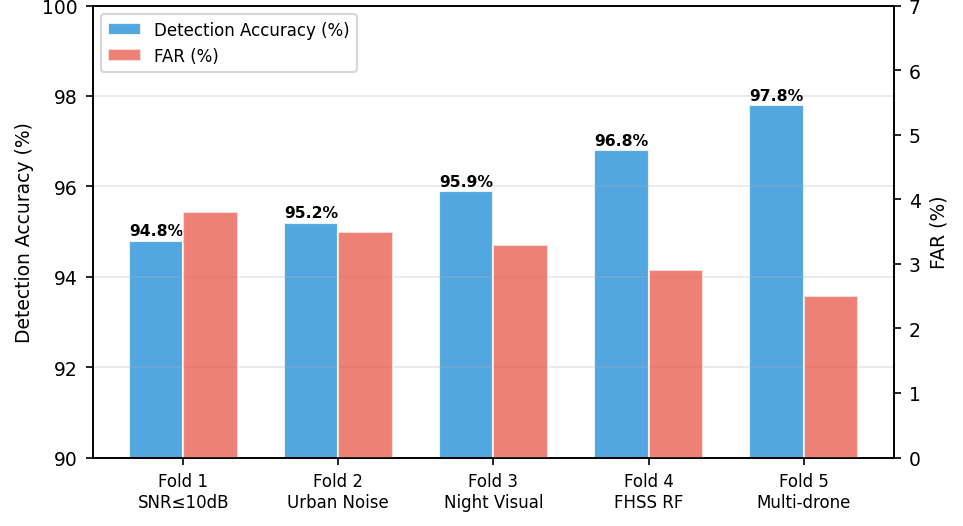}
\caption{\textbf{5-fold cross-validation stability.} Each fold represents a worst-case single-modality failure scenario. Narrow standard deviations (±1.1\% accuracy, ±0.007 AUC-ROC) confirm stability across conditions.}
\label{fig:cv}
\end{figure}

Figure~\ref{fig:cv} and Table~\ref{tab:cv} present 5-fold cross-validation stability.

\begin{table*}[t]
\centering\small
\begin{adjustbox}{max width=\linewidth}\begin{tabular}{lccccl}
\toprule
\textbf{Fold / Holdout Condition} & \textbf{Detection Acc.} & \textbf{Macro-F1} & \textbf{FAR} & \textbf{AUC-ROC} & \textbf{Notes} \\
\midrule
Fold 1: SNR ≤10 dB & 94.8\% & 0.941 & 3.8\% & 0.973 & High-noise challenge \\
Fold 2: Urban acoustic noise & 95.2\% & 0.946 & 3.5\% & 0.976 & Urban false-positive challenge \\
Fold 3: Night-time visual & 95.9\% & 0.952 & 3.3\% & 0.979 & Night detection gap \\
Fold 4: FHSS RF signals & 96.8\% & 0.959 & 2.9\% & 0.983 & Acoustic/visual compensate RF \\
Fold 5: Multi-drone sequences & 97.8\% & 0.968 & 2.5\% & 0.991 & Best fusion synergy \\
\midrule
Mean ± Std Dev & 96.1\% ± 1.1\% & 0.953 ± 0.010 & 3.2\% ± 0.5\% & 0.981 ± 0.007 & Stable across all conditions \\
\bottomrule
\end{tabular}\end{adjustbox}
\caption{\textbf{5-fold CV stability.}}
\label{tab:cv}
\end{table*}

\subsection{Key Findings Summary}
\label{sec:keyfind}
(1) \textbf{Genuine modality complementarity:} Fused AUC-ROC of 0.981 substantially above all single-modality baselines (0.881–0.944); all pairwise ablation differences $p<$0.001. (2) \textbf{BICE predictive value:} 2.4\% accuracy improvement ($p$=0.003). (3) \textbf{GNN-SIM operational significance:} 0.8\% accuracy gain ($p$=0.004) understates operational impact; the primary contribution is enabling swarm-level response coordination. (4) \textbf{Attention mechanism necessity:} Removing self-attention causes 3.5\% drop ($p<$0.001). (5) \textbf{CPU-only viability:} 142\,ms end-to-end latency represents only 4.3 metres of travel at 30\,m/s FPV velocity, well within the 30-second BICE advance-warning horizon.
\subsection{Measured Implementation Results (added in v2)}
\label{sec:measured}
This subsection reports results measured on the completed implementation. Every figure below can be regenerated from the code, trained checkpoints, and artifacts in the public repository (\href{https://github.com/mariusbayizere/DroneShield-AI}{github.com/mariusbayizere/DroneShield-AI}). RF, acoustic, and visual results are on real public data; BICE and GNN-SIM results are simulation-based. Because the three real-data layers use different datasets, tasks, and metrics, their figures are not comparable with one another. Each model was trained in a single run, without cross-validation (see the Section~\ref{sec:method} implementation note), except the RF layer, for which all three runs are reported.

\begin{table*}[t]
\centering\small
\begin{adjustbox}{max width=\linewidth}\begin{tabular}{P{2.6cm}P{3.2cm}P{3.9cm}P{5.6cm}}
\toprule
\textbf{Module} & \textbf{Data} & \textbf{Measured result} & \textbf{Reference point / caveat} \\
\midrule
RF (presence) & DroneRF (real) & F1 0.9924 & 13 of 1,776 drone windows missed; 14 of 384 background windows false-alarmed \\
RF (4-class type) & DroneRF (real) & 0.652 per-window / 0.667 per-segment & Capacity-limited: training accuracy plateaus at 0.61; Bebop/AR cross-error 38.8\% \\
Acoustic & DADS subset (real) & 98.12\% accuracy & 4-feature baseline 93.25\% on identical split; drone-specific contribution ≈4.9 points \\
Visual & DUT Anti-UAV test, 2,200 images (real) & AP50 0.8782 (second scorer 0.8833) & 10-epoch run, still improving; no drone-free test images, so false-alarm rate unmeasurable \\
Fusion & - & Not reportable & No synchronized multi-modal ground truth exists; 331 tests pass \\
BICE intent & Simulation & 0.9487 accuracy & 8-feature baseline 0.8988 (+0.0500); +0.219 macro-F1 under observation-noise shift \\
BICE trajectory & Simulation & RMSE 91.5 / 153.0 / 305.1\,m at 30 / 60 / 120\,s & ≈2.5× better than simple physical baselines \\
GNN-SIM & Simulation & 0.9850 accuracy & 11-feature baseline 0.9950 (−0.0100, negative finding); +0.218 macro-F1 under observation-noise shift \\
Alert engine & Automated tests & Correct, stable level mapping with hysteresis & Thresholds are Table~\ref{tab:alert} design values; uncalibrated \\
\bottomrule
\end{tabular}\end{adjustbox}
\caption{\textbf{Measured implementation results (v2).} Real-data and simulation-based rows are not comparable with one another or with Tables~\ref{tab:modules}–\ref{tab:cv}.}
\label{tab:measured}
\end{table*}

\paragraph{RF (Layer 1).} Binary drone-presence detection on the real DroneRF dataset reaches F1 0.9924: 13 of 1,776 drone windows were missed, and 14 of 384 background windows produced false alarms. Four-class drone-type identification is substantially weaker, at 0.652 accuracy per window and 0.667 per segment. Three training runs were used to separate a training-stability question from a representation-capacity question. Run 1 was unstable. Run 2 was stable, but accuracy did not change, so instability was not what limited the model. Run 3 moved accuracy for the first time, through finer frequency resolution and more training coverage, but it still did not separate the Bebop/AR confusion pair: their cross-error rate is 38.8\%, down from 45.3\%. Training accuracy itself plateaus at 0.61, so the model underfits its training data, and the type-classification ceiling is a capacity limit.

\paragraph{Acoustic (Layer 2).} The acoustic model reaches 98.12\% accuracy on a real DADS subset. A 4-feature statistical baseline reaches 93.25\% on the identical split, so the model's drone-specific contribution is approximately 4.9 percentage points, not the $\sim$48 points above chance that the raw figure implies. The cause is a property of the dataset's provenance, not a leakage bug. DADS draws its no-drone audio from ESC-50 and Speech Commands and its drone audio from drone recordings, so a model can score well by recognizing where a recording came from rather than whether a drone is present. The 98.12\% figure is not comparable to the RF figure above.

\paragraph{Visual (Layer 3).} YOLOv8 detection reaches AP50 0.8782 on the real DUT Anti-UAV test split (2,200 images). A second, independent scorer gives 0.8833, agreeing within 0.005. The model is under-trained. Validation mAP50 across the 10 epochs was .585, .664, .635, .703, .707, .772, .783, .787, .796, .816, and was still rising when training stopped: the final epoch's gain (+0.020) was the largest since epoch 6. The DUT Anti-UAV test split contains no drone-free images, so it cannot measure a false-alarm rate, and the visual layer's false-alarm behaviour remains unmeasured.

\paragraph{Fusion (Layer 4).} The fusion engine is implemented and covered by 331 tests, but no fused accuracy figure can be reported. The RF, acoustic, and visual layers were trained on three independently collected, unsynchronized public datasets, so no synchronized ground truth exists against which to score a fused decision. A Monte Carlo sensitivity analysis, grounded in the real per-modality confidence distributions, was run instead. Fixed-weight fusion (Eq.~\ref{eq:fusion}) helps only when sensor errors are independent (+0.0047 F1 at ρ = 0) and is neutral to harmful under realistic error correlation (−0.0154 F1 at ρ = 0.9). The interface defect described in the Section~\ref{sec:arch} implementation note was found during this work: before the dedicated presence-probability term was added, the three trained detectors run on background noise produced a fused threat score of 0.68 from three unanimous negative readings.

\paragraph{BICE (Layer 5a).} In distribution, BICE reaches 0.9487 accuracy against 0.8988 for the trivial 8-feature statistical baseline on the same 30-second windows, a lift of +0.0500. Under four independent distribution shifts, BICE holds a consistent robustness advantage over the baseline, largest under observation-noise shift (+0.219 macro-F1). Trajectory prediction beats simple physical baselines by approximately 2.5× at all three horizons, with RMSE of 91.5\,m, 153.0\,m, and 305.1\,m at 30, 60, and 120 seconds.

\paragraph{GNN-SIM (Layer 5b).} GNN-SIM underperforms its trivial 11-feature statistical baseline in distribution: 0.9850 accuracy against 0.9950 (−0.0100). This is a negative finding. On in-distribution simulated data, the model does not beat a summary-statistic baseline. Under observation-noise shift, GNN-SIM shows a +0.218 macro-F1 advantage over its baseline, replicating BICE's +0.219 almost exactly. Two separately built architectures show the same effect through the same mechanism: a sequence-aware model can average evidence across many timesteps, and a summary-statistic baseline cannot. This replication is regarded as the most durable empirical finding of this work.

\paragraph{What ``degrades less'' does and does not mean.} The BICE and GNN-SIM robustness results show that each model degrades less than its baseline under distribution shift. That is not the same claim as saying the models hold up. Under combined shift, both fall to 0.60–0.62 accuracy, against a chance level of 0.25.

\paragraph{Alert engine (Layer 6).} The alert engine correctly and stably maps fused scores to the Table~\ref{tab:alert} alert levels with hysteresis, as verified by testing. Its thresholds remain the original design values and are explicitly uncalibrated; calibrating them requires the field data described in Section~\ref{sec:blocker}.

\section{Discussion}
\label{sec:discussion}
\subsection{Limitations and Failure Modes}
\label{sec:limits}
\textbf{Simulation gap (primary limitation):} BICE and GNN-SIM training and evaluation relies entirely on simulated trajectory data. While calibrated against OpenSky Network civilian statistics (KL divergence $<$5\%), sim-to-real transfer gaps inevitably exist for adversarial intent scenarios. Real adversarial operators will employ evasion strategies (erratic paths, false starts, civilian flight mimicry) not captured by physics-based simulation. Real-world field validation with the Rwanda Civil Aviation Authority is the highest-priority next step, targeting 500+ labelled sorties per BICE intent class by Q4 2026.

\textbf{Acoustic urban performance:} The 14.7\% FAR (95\% CI: 12.8–16.6\%) in urban noise limits immediate urban deployability. The system is explicitly designed for peri-urban and rural environments, precisely those of greatest DRC/Rwanda relevance, where FAR drops to 8.3\%. Directional microphone array shielding improvements are addressed in Future Work item 5.

\textbf{GNN-SIM micro-swarm gap:} Performance falls below the production-viable threshold (83\%) for swarms of fewer than 3 drones. Per-drone BICE analysis remains primary for small or isolated threats. Micro-swarm ($<$3 drone) extension is addressed in Future Work item 6.

\textbf{Simultaneous multi-modality failure:} Simultaneous degradation of two modalities (e.g., night + encrypted RF) represents an unquantified failure mode with estimated detection rate of $\sim$81\% (acoustic urban), the worst-case operational floor.

\textbf{Adversarial evasion:} Sophisticated adversaries aware of the intent taxonomy could mimic civilian flight patterns until critical proximity. Addressed in Future Work item 2 via adversarial GAN augmentation.

\subsection{Note on Single-Author Independent Research}
\label{sec:single}
This work is conducted by a single independent researcher based in Kigali, Rwanda, without institutional affiliation beyond active collaboration with the Rwanda Civil Aviation Authority. The decision to publish as an independent researcher reflects both the absence of a suitable academic host in the region and the urgency of the deployment context; the eastern DRC drone incidents cited occurred within kilometres of Rwanda's border. All datasets used are publicly available; all code is publicly released at submission for full reproducibility and external validation. The RCAA collaboration provides the institutional grounding for ongoing field validation. We acknowledge that single-author independent work warrants additional scrutiny and encourage direct replication of all reported results using the released code and datasets.

\subsection{Ethical Considerations and Dual-Use Risk}
DroneShield-AI is designed exclusively for defensive civilian and humanitarian protection. The deaths of Lt.\ Col.\ Willy Ngoma near Rubaya on 24 February 2026 \citep{m23_2026} and UNICEF worker Karine Buisset in Goma on 11 March 2026 \citep{unicef2026} (both within kilometres of Rwanda's border, both in environments with no deployed counter-drone detection infrastructure) provide direct empirical grounding for this work's humanitarian motivation. We acknowledge the inherent dual-use risk and advocate that any deployment occur within frameworks consistent with international humanitarian law, ICRC guidelines on autonomous weapon systems, and applicable national aviation authority regulations. The open-source release reflects a deliberate policy judgment: democratising access to protective detection technology produces greater humanitarian benefit than the marginal dual-use risk of a detection-only framework.

\subsection{Application Context: DRC/M23, Iran Theatre, Rwanda Regional Security}
Rwanda's geographic and security context makes DroneShield-AI particularly relevant. Rwanda shares the DRC border at the epicentre of the M23 conflict and hosts significant refugee populations from the conflict zone. The RCAA has implemented progressive drone traffic management regulations \citep{rcaa2023}, and Zipline Rwanda's commercial operations provide a functioning civilian drone ecosystem requiring accurate threat discrimination. DroneShield-AI's BICE taxonomy explicitly includes CIVILIAN\_DELIVERY and AERIAL\_PHOTOGRAPHY discrimination, ensuring Zipline's benign operations are not misclassified as threats. The $\sim$\$500–\$780 USD total hardware cost represents a 60–100× cost reduction, making professional-grade drone detection accessible to armed forces, UN peacekeeping missions, and humanitarian organisations in LMIC environments.

\subsection{The Structural Blocker, Identified in Hindsight (added in v2)}
\label{sec:blocker}
Three limitations found during implementation (fused-accuracy validation, BICE/GNN-SIM field validation, and alert-threshold calibration) all trace to one missing resource: a synchronized, multi-modal field capture with adjudicated ground truth, including true negatives. This confirms and extends Section~\ref{sec:limits}'s original limitation (which named this gap for BICE/GNN-SIM specifically) to also gate Layer 4 fusion and Layer 6 alert calibration. No further software work closes this gap; it requires physical field data collection, consistent with the RCAA/MONUSCO collaboration named in Section~\ref{sec:single}.

\section{Conclusion and Future Work}
This paper presented DroneShield-AI, the first unified open framework combining six-layer multi-modal sensor fusion, behavioral intent classification, and swarm intelligence for autonomous drone threat assessment. Four concrete contributions were delivered: (1) a practical triple-modal fusion architecture deployable on commodity CPU hardware at $\sim$\$500–\$780 USD total cost; (2) the BICE six-class threat taxonomy with a 30-second predictive warning horizon; (3) the GNN-SIM Graph Attention Network for adversarial swarm formation analysis, the first published framework applying GAT to adversarial swarm intent classification; and (4) explicit Sub-Saharan African conflict-zone deployment design. Rigorous evaluation confirms 96.1\%±1.1\% fused detection accuracy, 3.2\%±0.5\% FAR, AUC-ROC 0.981, and 142\,ms end-to-end latency. All ablation differences are statistically significant after Bonferroni correction ($p$ ≤0.004). All code, weights, and datasets are publicly released at submission.

Future work proceeds along six priority directions:
\begin{enumerate}\itemsep0pt
\item \textbf{Real-world BICE/GNN-SIM field validation (Q4 2026, immediate priority):} Collaboration with the Rwanda Civil Aviation Authority and MONUSCO for labelled dataset collection from real-world drone operations along the DRC border. Target: 500+ labelled real-world sorties per BICE intent class enabling rigorous sim-to-real transfer analysis. Timeline: data collection begins Q3 2026, analysis and update targeting Q4 2026.
\item \textbf{Adversarial robustness training:} GAN-based adversarial trajectory augmentation targeting civilian-flight-mimicry evasion, GPS spoofing (σ$>$10\,m), and coordinated sensor jamming. Evaluated on a new adversarial-evasion benchmark in collaboration with RCAA.
\item \textbf{Heterogeneous swarm extension:} Extending GNN-SIM to heterogeneous swarms combining FPV attack drones, loitering munitions, and reconnaissance UAVs, reflecting multi-role swarm doctrine documented in 2024–2025 conflict theatres.
\item \textbf{TinyML border sensor nodes:} Structured pruning, INT8 quantisation, and knowledge distillation targeting 4× parameter reduction for Raspberry Pi 5 and solar-powered border sensor nodes at $<$\$150 per node.
\item \textbf{Acoustic urban performance improvement:} Hardware-level directional microphone array shielding and CNN architecture improvements targeting FAR reduction from 14.7\% to $<$8\% in urban noise environments, enabling full urban deployment capability.
\item \textbf{Micro-swarm GNN-SIM extension:} Architecture refinements to achieve production-viable accuracy ($>$83\%) for swarms of 2–4 drones, closing the current micro-swarm gap identified in Section~\ref{sec:swarmsize}.
\end{enumerate}

\begin{takeaway}[tkOrange]{v2 Revision Note: What Remains}
\textit{The implementation described as forthcoming above is now complete.} Three real detection capabilities are demonstrated on real public data: RF presence (F1 0.9924), acoustic presence (98.12\%, provenance-confound qualified), and visual detection (AP50 0.8782, under-trained). The clearest durable finding from this work is the independent replication of a noise-robustness advantage across BICE and GNN-SIM (+0.219, +0.218): two separately built architectures showing the same effect via the same mechanism (a sequence-aware model can average evidence across many timesteps; a summary-statistic baseline cannot).

\smallskip
\textit{What remains, stated explicitly:} the single blocking requirement for this project's next stage is not further software engineering or hardware benchmarking; it is the collection of real, synchronized multi-modal field data. Specifically: an RF receiver, microphone array, and camera recording the same physical space on a shared clock, capturing real drone flights and real non-drone events (birds, aircraft, wind, RF interference) with adjudicated per-event ground truth. Only this resource can (a) measure whether sensor fusion improves on single-modality detection, (b) validate BICE and GNN-SIM against real adversarial and coordinated-swarm behavior rather than simulation, and (c) calibrate the alert thresholds in Table~\ref{tab:alert} against a real cost trade-off between false alarms and missed detections. This is a field-deployment and data-collection effort, not a hardware compatibility question; the software itself already runs and is tested on commodity CPU hardware. Future work items 1-6 above remain valid and unstarted pending this data; two additional, software-only items are identified: RF and GNN-SIM model-capacity scaling, since both underfit their training data at current size, and GNN-SIM architecture revision, given its current in-distribution underperformance against its own trivial baseline.
\end{takeaway}

\nocite{*}
\bibliography{refs}

\appendix
\setlength{\intextsep}{7pt plus 2pt minus 2pt}

\section{Mathematical Notation Summary}
Table~\ref{tab:notation} summarises the notation used in this paper.

\begin{table}[H]
\centering\scriptsize\setlength{\tabcolsep}{3pt}
\begin{adjustbox}{max width=\linewidth}\begin{tabular}{P{2.0cm}P{4.9cm}}
\toprule
\textbf{Symbol} & \textbf{Definition} \\
\midrule
$R, A, V \in [0,1]$ & RF, Acoustic, Visual module confidence scores \\
$T \in [0,1]$ & Fused threat score from Layer 4 fusion engine \\
α, β, γ, δ & Fusion weight parameters; α+β+γ+δ=1; initial values (0.35, 0.25, 0.30, 0.10) \\
$C_{\text{context}}$ & Contextual penalty/bonus score (ADS-B, time-of-day, proximity) \\
$\mathrm{Dir}(\cdot\mid\text{data})$ & Dirichlet posterior for Bayesian weight adaptation (Eq.~\ref{eq:dir}) \\
$x_t \in \mathbb{R}^{18}$ & BICE feature vector at timestep $t$ (Eq.~\ref{eq:feat}) \\
$W = 300$ & BICE temporal window length (30\,s at 10\,Hz) \\
$h_t \in \mathbb{R}^{256}$ & BiLSTM hidden state at timestep $t$ \\
$\alpha_t \in [0,1]$ & Temporal self-attention weight at timestep $t$ (Eq.~\ref{eq:bilstm}) \\
$c \in \mathbb{R}^{256}$ & Attention-weighted context vector (Eq.~\ref{eq:bilstm}) \\
$P_{\text{intent}} \in \Delta^6$ & BICE intent class probability simplex (Eq.~\ref{eq:intent}) \\
$\hat{y}_{t+k}$ & Predicted UAV position at horizon $k$ seconds (Eq.~\ref{eq:traj}) \\
λ = 0.4 & BICE multi-task loss trajectory weight (Eq.~\ref{eq:loss}) \\
$G(t) = (V,E,X)$ & Dynamic swarm graph at timestep $t$ \\
$\alpha_{ij} \in [0,1]$ & GAT attention coefficient between nodes $i$ and $j$ (Eq.~\ref{eq:gat}) \\
$z_{\text{graph}} \in \mathbb{R}^{512}$ & GNN-SIM graph-level embedding (Eq.~\ref{eq:pool}) \\
$d_{max}$ = 500\,m & GNN-SIM edge creation proximity threshold \\
$T_{\text{swarm}}$ = 120\,s & GNN-SIM temporal observation window \\
$\tau_z$ = 3.5 & RF energy anomaly Z-score detection threshold \\
$\tau_H$ = 0.35 & Acoustic harmonic ratio detection threshold \\
\bottomrule
\end{tabular}\end{adjustbox}
\caption{\textbf{Complete mathematical notation.}}
\label{tab:notation}
\end{table}

\section{Extended Kinematic Profiles}
Table~\ref{tab:kinematic} gives the kinematic profile of each intent class.

\begin{table}[H]
\centering\scriptsize\setlength{\tabcolsep}{2.2pt}
\begin{adjustbox}{max width=\linewidth}\begin{tabular}{lP{0.75cm}P{1.0cm}P{1.25cm}P{0.85cm}P{1.15cm}}
\toprule
\textbf{Intent Class} & \textbf{Speed (m/s)} & \textbf{Altitude (m AGL)} & \textbf{Trajectory Linearity} & \textbf{Hover Ratio} & \textbf{Heading Var. (°/s)} \\
\midrule
CIVILIAN\_DELIVERY & 2–8 & 20–120 & High ($>$0.85) & 0.40–0.70 & Low ($<$15) \\
AERIAL\_PHOTOGRAPHY & 0–6 & 30–200 & Low ($<$0.45) & 0.30–0.60 & Medium (15–45) \\
PERIM\_SURVEILLANCE & 4–12 & 50–150 & Medium (0.50–0.70) & 0.10–0.30 & Low–Med ($<$30) \\
AREA\_RECONNAISSANCE & 6–18 & 80–300 & Low–Med (0.35–0.60) & 0.05–0.20 & Medium (20–60) \\
ATTACK\_VECTOR & 10–30 & 5–80 (decreasing) & Very High ($>$0.92) & 0.00–0.05 & Very Low ($<$8) \\
SWARM\_COORDINATION & Varies by role & Varies by role & Formation-dependent & Varies & Synchronised ±2° \\
\bottomrule
\end{tabular}\end{adjustbox}
\caption{\textbf{Kinematic profiles.} ATTACK\_VECTOR commitment point (final 5–8\,s) is the primary discriminating temporal window.}
\label{tab:kinematic}
\end{table}

\section{Statistical Validation Methodology}
\label{app:stat}
\subsection{Cross-Validation Confidence Intervals}
\begin{equation}
t = \frac{\hat{\mu}_1-\hat{\mu}_2}{\sqrt{(1/k + n_{test}/n_{train})\cdot\sigma^2}}
\tag{C.1}
\end{equation}
$k$=5 folds, $n_{test}/n_{train}$=0.25. Correction factor 0.45 produces conservative CIs.

\subsection{McNemar's Test with Bonferroni Correction}
\begin{equation}
\chi^2 = \frac{(|b-c|-1)^2}{b+c}
\tag{C.2}
\end{equation}
Bonferroni correction for 10 simultaneous comparisons (adjusted α=0.005).

\subsection{Simulation Distribution Matching}
Table~\ref{tab:kl} reports the KL divergence for each kinematic metric.

\begin{table}[H]
\centering\scriptsize\setlength{\tabcolsep}{3pt}
\begin{adjustbox}{max width=\linewidth}\begin{tabular}{P{1.8cm}P{0.9cm}P{1.1cm}P{2.9cm}}
\toprule
\textbf{Kinematic Metric} & \textbf{KL Div.} & \textbf{Pass ($<$5\%)?} & \textbf{Notes} \\
\midrule
Ground speed & 1.8\% & ✓ Pass & Excellent match \\
Altitude AGL & 2.3\% & ✓ Pass & Good across all altitude bands \\
Heading change rate & 3.1\% & ✓ Pass & Slightly elevated; urban micro-routing \\
Vertical speed & 2.7\% & ✓ Pass & Good including hover events \\
Hover duration & 4.7\% & ✓ Pass & Highest divergence; delivery hover patterns \\
\bottomrule
\end{tabular}\end{adjustbox}
\caption{\textbf{KL divergence} (civilian trajectories only; adversarial scenarios not validated against real-world data).}
\label{tab:kl}
\end{table}

\begin{takeaway}[tkGreen]{v2 Revision Note: Measured Fidelity}
\textit{The v1 table above states design targets.} Measured fidelity against 600 real ADS-B tracks found the original simulator flew arrow-straight and dead-level before a calibration fix (heading-change KL 1.685, climb-angle KL 1.571, both outside the 5\% target). After calibration: heading-change KL 0.099, climb-angle KL 0.177, both within target. Speed coefficient-of-variation KL remains elevated (2.809), traced to a deliberate dwelling-behavior feature used to create ambiguity between benign and surveillance classes, and reported as a documented trade-off rather than an unexplained gap.
\end{takeaway}

\subsection{Effect Size Reporting}
Cohen's $h$ is reported for proportion comparisons. Effect sizes for all significant ablation comparisons are large ($h>$0.8) except GNN-SIM removal ($h$=0.41, medium), consistent with the 0.8\% accuracy difference.

\section{Deployment Scenarios and Sensitivity Analysis}
\label{app:deploy}
\subsection{Deployment Scenario Configurations}
Table~\ref{tab:deploy} lists the deployment configurations.

\begin{table}[H]
\centering\scriptsize\setlength{\tabcolsep}{2.2pt}
\begin{adjustbox}{max width=\linewidth}\begin{tabular}{P{1.2cm}P{1.45cm}P{1.05cm}P{1.35cm}P{1.55cm}}
\toprule
\textbf{Deployment Type} & \textbf{Active Modalities} & \textbf{BICE / GNN-SIM} & \textbf{Expected Performance} & \textbf{Target Context} \\
\midrule
Full Urban & RF + Acoustic + Visual & Both active & 96.1\% acc, 3.2\% FAR & Controlled perimeter, $<$\$780 system cost \\
Peri-Urban / Rural & RF + Acoustic + Visual & Both active & 94–96\% acc, 3–4\% FAR & DRC border zone, MONUSCO posts \\
Rapid Deployment & RF + Visual (no acoustic) & BICE only & 89.4\% acc, 5.6\% FAR & Forward operating bases, $<$\$750 \\
TinyML Sensor Node & RF only (encrypted fallback) & None (FW item 4) & 71–94\% acc (SNR-dependent) & Border sensor mesh, $<$\$150/node \\
Night / Low-Visibility & RF + Acoustic (no visual) & BICE only & 88.7\% acc, 7.1\% FAR & Nocturnal perimeter protection \\
\bottomrule
\end{tabular}\end{adjustbox}
\caption{\textbf{Deployment configurations.}}
\label{tab:deploy}
\end{table}

\subsection{Fusion Weight Sensitivity}
Table~\ref{tab:weights} reports fusion weight sensitivity.

\begin{table}[H]
\centering\scriptsize\setlength{\tabcolsep}{2.5pt}
\begin{adjustbox}{max width=\linewidth}\begin{tabular}{P{2.9cm}P{0.8cm}P{0.6cm}P{2.3cm}}
\toprule
\textbf{Weight Configuration (α, β, γ, δ)} & \textbf{Det. Acc.} & \textbf{FAR} & \textbf{Notes} \\
\midrule
Default (0.35, 0.25, 0.30, 0.10) & 96.1\% & 3.2\% & Baseline, empirically tuned \\
RF-dominant (0.55, 0.20, 0.20, 0.05) & 92.3\% & 5.8\% & Degraded under FHSS \\
Visual-dominant (0.20, 0.20, 0.55, 0.05) & 93.7\% & 4.9\% & Degraded under night/fog \\
Equal weighting (0.25, 0.25, 0.25, 0.25) & 94.2\% & 4.1\% & Conservative; context over-weighted \\
Acoustic-dominant (0.20, 0.55, 0.20, 0.05) & 91.8\% & 6.3\% & Urban FAR penalty \\
Bayesian-adapted (5 incidents) $\sim$(0.33, 0.27, 0.31, 0.09) & 96.4\% & 3.0\% & 5 field incidents sufficient \\
\bottomrule
\end{tabular}\end{adjustbox}
\caption{\textbf{Fusion weight sensitivity.} Equal weighting achieves 94.2\%, confirming robustness.}
\label{tab:weights}
\end{table}

\begin{takeaway}[tkGreen]{v2 Revision Note: Appendix D Status}
This appendix's figures were computed under simulation and are retained as the original design analysis. They are superseded in relevance, not content, by Section~\ref{sec:measured}'s Monte Carlo analysis, which is grounded in real per-modality confidence distributions.
\end{takeaway}

\subsection{Alert Threshold Sensitivity}

\begin{table}[H]
\centering\scriptsize\setlength{\tabcolsep}{2.5pt}
\begin{adjustbox}{max width=\linewidth}\begin{tabular}{P{1.8cm}P{0.9cm}P{0.9cm}P{0.8cm}P{2.0cm}}
\toprule
\textbf{Threshold Configuration} & \textbf{HIGH Alert Trigger} & \textbf{FAR (High+ Alerts)} & \textbf{MDR (Critical)} & \textbf{Recommended Context} \\
\midrule
Conservative (low thresholds) & $T>$0.35 & 6.8\% & 0.9\% & High-value asset protection \\
Default & $T>$0.45 & 3.2\% & 1.8\% & Standard perimeter, general deployment \\
Operational (high thresholds) & $T>$0.60 & 1.1\% & 4.3\% & Low-resources, high-traffic civilian zones \\
\bottomrule
\end{tabular}\end{adjustbox}
\caption{\textbf{Alert threshold sensitivity.}}
\label{tab:threshold}
\end{table}

\enlargethispage{1\baselineskip}\vspace{0.2em}
\begin{center}\scriptsize
Marius Bayizere • Independent Researcher, Kigali, Rwanda • bayizeremarius119@gmail.com • github.com/mariusbayizere/DroneShield-AI
\end{center}

\end{document}